\documentclass[11pt,a4paper]{article}
\usepackage[hyperref]{acl2019}
\usepackage{times}
\usepackage{latexsym}
\usepackage{booktabs}
\usepackage{graphicx}
\usepackage{multirow}

\usepackage{subfigure}
\usepackage[skip=5pt]{caption}
\usepackage{url}

\aclfinalcopy 
\def\aclpaperid{938} 

\title{Better Character Language Modeling Through Morphology}
\author{Terra Blevins$^{1}$ and Luke Zettlemoyer$^{1, 2}$ \\
        $^1$Paul G. Allen School of Computer Science \& Engineering, University of Washington \\
        $^2$Facebook AI Research \\
        {\tt \{blvns, lsz\}@cs.washington.edu}}

\date{}

\begin{document}
\maketitle
\begin{abstract}
We incorporate morphological supervision into character language models (CLMs) via multitasking and show that this addition improves bits-per-character (BPC) performance across 24 languages, even when the morphology data and language modeling data are disjoint. Analyzing the CLMs shows that inflected words benefit more from explicitly modeling morphology than uninflected words, and that morphological supervision improves performance even as the amount of language modeling data grows. We then transfer morphological supervision across languages to improve language modeling performance in the low-resource setting.
\end{abstract}

\section{Introduction}

Character language models (CLMs) are distributions over sequences of characters \cite{sutskever2011}, in contrast to traditional language models which are distributions over sequences of words. CLMs eliminate the need for a fixed word vocabulary, and modeling text at the character level gives the CLM access to subword information. These attributes suggest that CLMs can model regularities that exist within words, such as morphological inflection. However, even large language modeling (LM) corpora have sparse coverage of inflected forms for morphologically-rich languages, which has been shown to make word and character language modeling more difficult \cite{gerz2018relation,cotterell2018}. Because to this, we hypothesize that accurately modeling morphology improves language modeling, but that it is difficult for CLMs to learn this from text alone.

Motivated by this hypothesis, we add morphology supervision to character language modeling and show that, across two benchmark datasets, multitasking morphology with CLMs improves bits-per-character (BPC) performance on twenty-four languages, even when the annotated morphology features and language modeling data do not overlap. We also show that models augmented by multitasking achieve better BPC improvements on inflected forms than on uninflected forms, and that increasing the amount of language modeling data does not diminish the gains from morphology. Furthermore, to augment morphology annotations in low-resource languages, we also transfer morphology information between pairs of high- and low-resource languages. In this cross-lingual setting, we see that morphology supervision from the high-resource language improves BPC performance on the low-resource language over both the low-resource multitask model and over adding language modeling data from the high-resource language alone. 

\section{Approach}

\begin{table*}[t]
    \resizebox{\textwidth}{!}{
    \begin{tabular}{ c c c c c c c c c c c c c c c}
    \toprule
    & \multicolumn{2}{c}{\textbf{CS}} & \multicolumn{2}{c}{\textbf{DE}} & \multicolumn{2}{c}{\textbf{EN}} & \multicolumn{2}{c}{\textbf{ES}} & \multicolumn{2}{c}{\textbf{FI}} & \multicolumn{2}{c}{\textbf{FR}} & \multicolumn{2}{c}{\textbf{RU}} \\ 
    \hline
    & dev & test & dev & test & dev & test & dev & test & dev & test & dev & test & dev & test \\
    \hline 
    \textbf{HCLM} & 2.010 & 1.984 & 1.605 & 1.588 & 1.591 & 1.538 & 1.548 & 1.498 & 1.754 & 1.711 & 1.499 &  1.467 & 1.777 & 1.761 \\
     \textbf{LM} & 2.013 & 1.972 & 1.557 & 1.538 & 1.543 & 1.488 & 1.571 & 1.505 & 1.725 & 1.699 & 1.357 & 1.305 & 1.745 & 1.724 \\
     \textbf{MTL} & \textbf{1.938} & \textbf{1.900} & \textbf{1.249} & \textbf{1.241} & \textbf{1.313} & \textbf{1.256} & \textbf{1.260} & \textbf{1.196} & \textbf{1.698} & \textbf{1.669} & \textbf{1.211} & \textbf{1.167} & \textbf{1.645} & \textbf{1.619} \\
    \hline
    $\Delta$ & 0.075 & 0.072 & 0.308 & 0.297 & 0.230 & 0.232
    & 0.311 & 0.309 & 0.027 & 0.030 & 0.146 & 0.138 & 0.100 & 0.105 \\
    \toprule
    \end{tabular}}
    \caption{Results on Multilingual Wikipedia Corpus (MWC) in bits per character (BPC). $\Delta$ calculated improvement in BPC from the baseline LM to MTL. HCLM is the best model from \citet{kawakami2017}.}
    \label{mwc-results-table}
\end{table*}
\paragraph{Language Modeling}
Given a sequence of characters $\textbf{c} = c_1, c_2,...,c_n$, our character-level language models calculate the probability of $\textbf{c}$ as 
\begin{equation}p(\textbf{c}) = \prod_{i=1}^{|\textbf{c}|} p(c_i | c_1, c_2,...,c_{i-1})\end{equation} 

Each distribution is an LSTM \cite{hochreiter1997} trained such that at each time step $t$, the model takes in a character $c_t$ and estimates the probability of the next character $c_{t+1}$ as
\begin{equation}p(c_{t+1}|c_{\leq t}) = g(\mathrm{LSTM}(\textbf{w}_t, \textbf{h}_{t-1}))\end{equation} 

\noindent
where $\textbf{h}_{t-1}$ is the previous hidden state of the LSTM, $\textbf{w}_t$ is the character embedding learned by the model for $c_t$, and $g$ is a softmax over the character vocabulary space.

We calculate the loss function of our language model $\mathcal{L}_{\mathrm{LM}}$ as the negative log-likelihood of the model on the character sequence $\textbf{c}$:
\begin{equation}\mathcal{L}_{\mathrm{LM}}(\textbf{c}) = \mathrm{NLL}(\textbf{c}) = -\sum_{i=1}^{|\textbf{c}|} \textrm{log }p(c_i | c_{<i})\end{equation} 

We then evaluate the trained model's performance with bits-per-character (BPC):
\begin{equation}\mathrm{BPC}(\textbf{c}) = -\frac{1}{|\textbf{c}|} \sum_{i=1}^{|\textbf{c}|} \textrm{log }p(c_i | c_{<i})\end{equation} 

\paragraph{Multitask Learning}
To add morphology features as supervision, we use a multitask learning (MTL) objective \cite{collobert2008} that combines loss functions for predicting different morphological tags with the language modeling objective.
Since morphological features are annotated at the word-level, we convert these annotations to the character level by placing each annotated word's tags as supervision on the first character (which we found to outperform supervising the last character in preliminary results).

This early placement allows the model to have access to the morphological features while decoding the rest of the characters in the word. Therefore, our morphology data $\textbf{m} = m_1, m_2, ..., m_n$ is a sequence of labeled pairs in the form $m_i = (x,y)$ where $x$ is a character and $y$ is a set of morphology tags for that character. For example, ``cats ran'' would be given to our model as the sequence (`c', Number=Pl), (`a', -), (`t', -), (`s', -), (` ', -), (`r', Tense=Past), (`a', -), (`n', -).

We modify the model's loss function to
\begin{equation}\mathcal{L}(\textbf{c}, \textbf{m}) = \mathcal{L}_{\mathrm{LM}}(\textbf{c}) + \delta \sum_{i=1}^{n} \mathcal{L}_i(\textbf{m})\end{equation} 
where $n$ is the number of morphological features we have annotated in a language, $\delta$ is a weighting parameter between the primary and auxiliary losses, $\mathcal{L}_{\mathrm{LM}}$ is the original language modeling loss, and $\mathcal{L}_i$ are the additional losses for each morphological feature (e.g., tense, number, etc). Because we include a separate loss for each morphological feature, each feature is predicted independently.

\section{Experimental Setup}
\paragraph{Datasets}
We obtain morphological annotations for 24 languages (Table \ref{ud-results-table}) from Universal Dependencies (UD; v.2.3), which consists of dependency parsing treebanks with morphology annotations on a large number of languages \cite{nivre2018}. These languages were chosen based on the size of their treebanks (to ensure a sufficient amount of morphology annotations); we also exclude languages that do not have morphology features annotated in the treebank.

For language modeling supervision, we train two sets of models. One set is trained with the text from the UD treebanks; the other set of models is trained on the Multilingual Wikipedia Corpus (MWC) \cite{kawakami2017}. This language modeling dataset consists of Wikipedia data across seven languages (Czech, German, English, Spanish, Finnish, French, and Russian). 

\paragraph{Model architecture}
Our models each consist of a stacked LSTM with 1024 hidden dimensions and a character embedding layer of 512 dimensions. We include two hidden layers in the language models trained on UD, and three hidden layers in those trained on MWC. The parameters that integrate multitasking into the model (the layer at which we multitask morphology and the weighting we give the morphology losses, $\delta$) are tuned individually for each language. Further hyperparameter and training details are given in the supplement. 

\begin{table}[t]
    \centering
    \small
    \begin{tabular}{ c c c | c c c}
        \toprule
        \textbf{Lang} & \textbf{ISO} & \textbf{\%Infl} & \textbf{LM} & \textbf{MTL} & \textbf{$\Delta$} \\
        \hline
        Bulgarian & BG & 39\% & 1.890 & 1.887 & 0.003 \\
        Catalan & CA & 31\% & 1.653 & 1.599 & 0.054 \\
        Czech & CS & 43\% & 2.045 & 1.832 & 0.213 \\
        Danish & DA & 30\% & 2.152 & 2.135 & 0.017 \\
        German & DE & 33\% & 1.917 & 1.881 & 0.036 \\
        English & EN & 15\% & 2.183 & 2.173 & 0.010 \\
        Spanish & ES & 28\% & 1.801 & 1.763 & 0.038 \\
        Farsi & FA & 27\% & 2.213 & 2.205 & 0.008 \\
        French & FR & 32\% & 1.751 & 1.712 & 0.039 \\
        Hindi & HI & 28\% & 1.819 & 1.773 & 0.046 \\
        Croatian & HR & 49\% & 1.866 & 1.841 & 0.025 \\
        Italian & IT & 36\% & 1.595 & 1.554 & 0.041 \\
        Latvian & LV & 47\% & 2.243 & 2.217 & 0.026 \\
        Dutch & NL & 19\% & 1.989 & 1.972 & 0.017 \\
        Polish & PL & 42\% & 2.218 & 2.154 & 0.064 \\
        Portuguese & PT & 31\% & 1.787 & 1.785 & 0.002 \\
        Romanian & RO & 42\% & 1.840 & 1.798 & 0.042 \\
        Russian & RU & 42\% & 1.993 & 1.824 & 0.169  \\
        Slovak & SK & 45\% & 2.705 & 2.686 & 0.019 \\
        Ukranian & UK & 40\% & 2.359 & 2.338 & 0.021 \\
        \hline 
        Estonian & ET & 49\% & 2.089 & 1.993 & 0.096 \\
        Finnish & FI & 55\% & 1.981 & 1.921 & 0.060 \\
        \hline
        Arabic & AR & 86\% & 1.724 & 1.708 & 0.016 \\
        Hebrew & HE & 42\% & 2.293 & 2.282 & 0.011 \\
        \toprule
    \end{tabular}
    \caption{BPC results on the Universal Dependencies (UD) test set. \%Infl is the inflection rate in each language.  Languages are grouped by fusional, agglutinative, and introflexive typologies, respectively.}
    \label{ud-results-table}
\end{table}

\begin{figure*}[t]
\subfigure[]{
    \centering
    \includegraphics[scale=0.35]{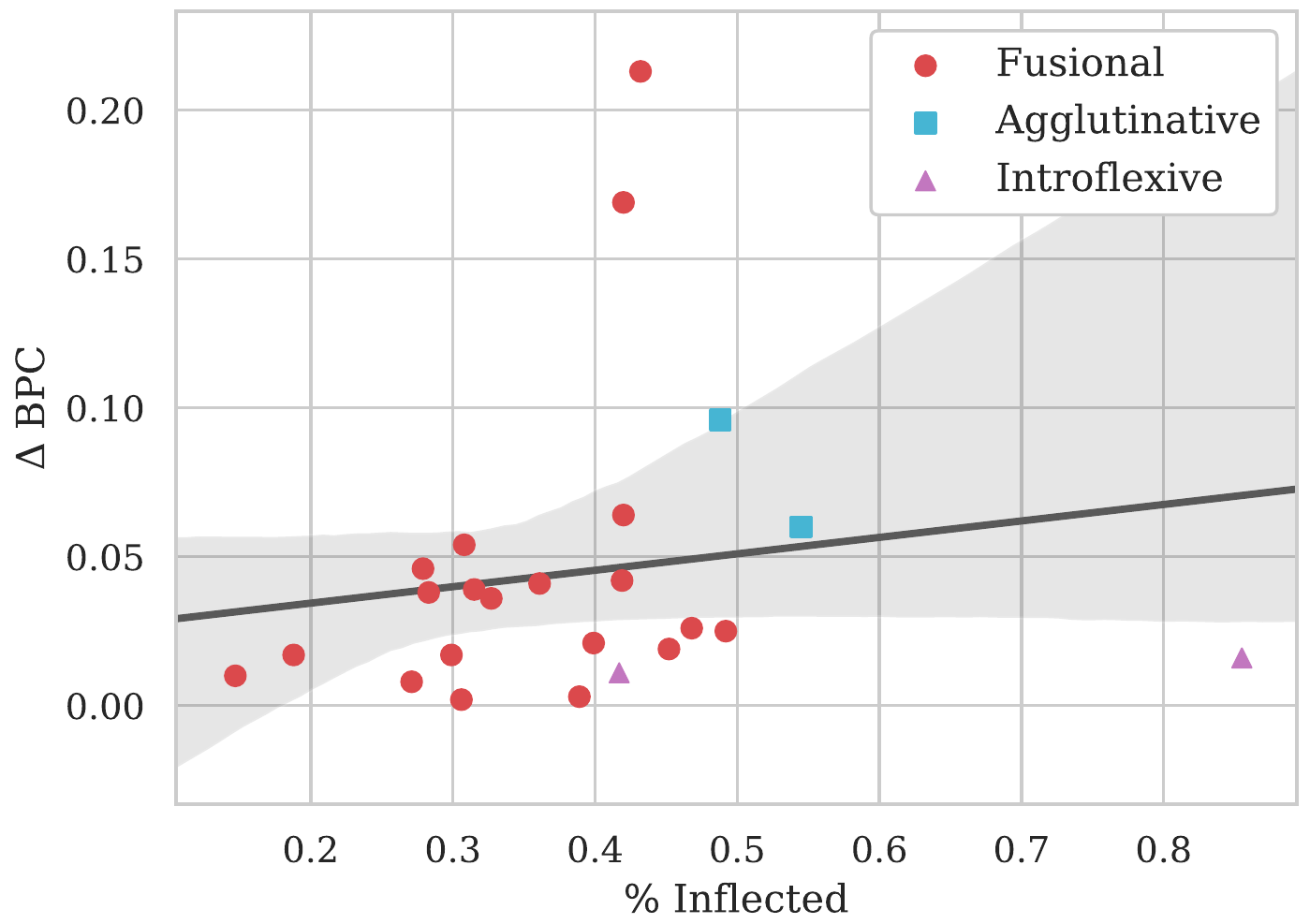}
    \label{ud-plot-infl}}
\subfigure[]{   
    \centering
    \includegraphics[scale=0.35]{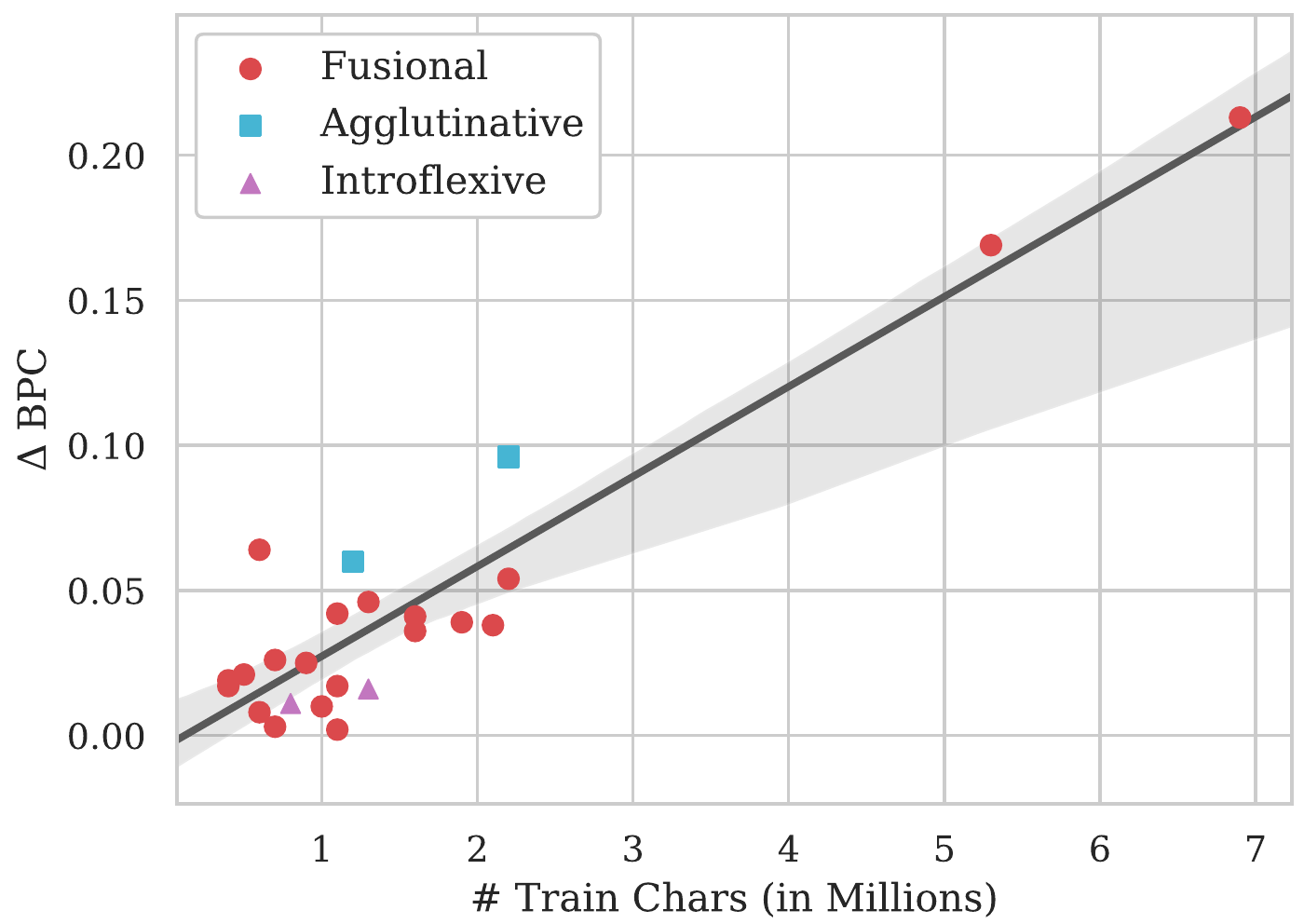}
    \label{ud-plot-train}}
\subfigure[]{ 
    \centering
    \includegraphics[scale=0.4]{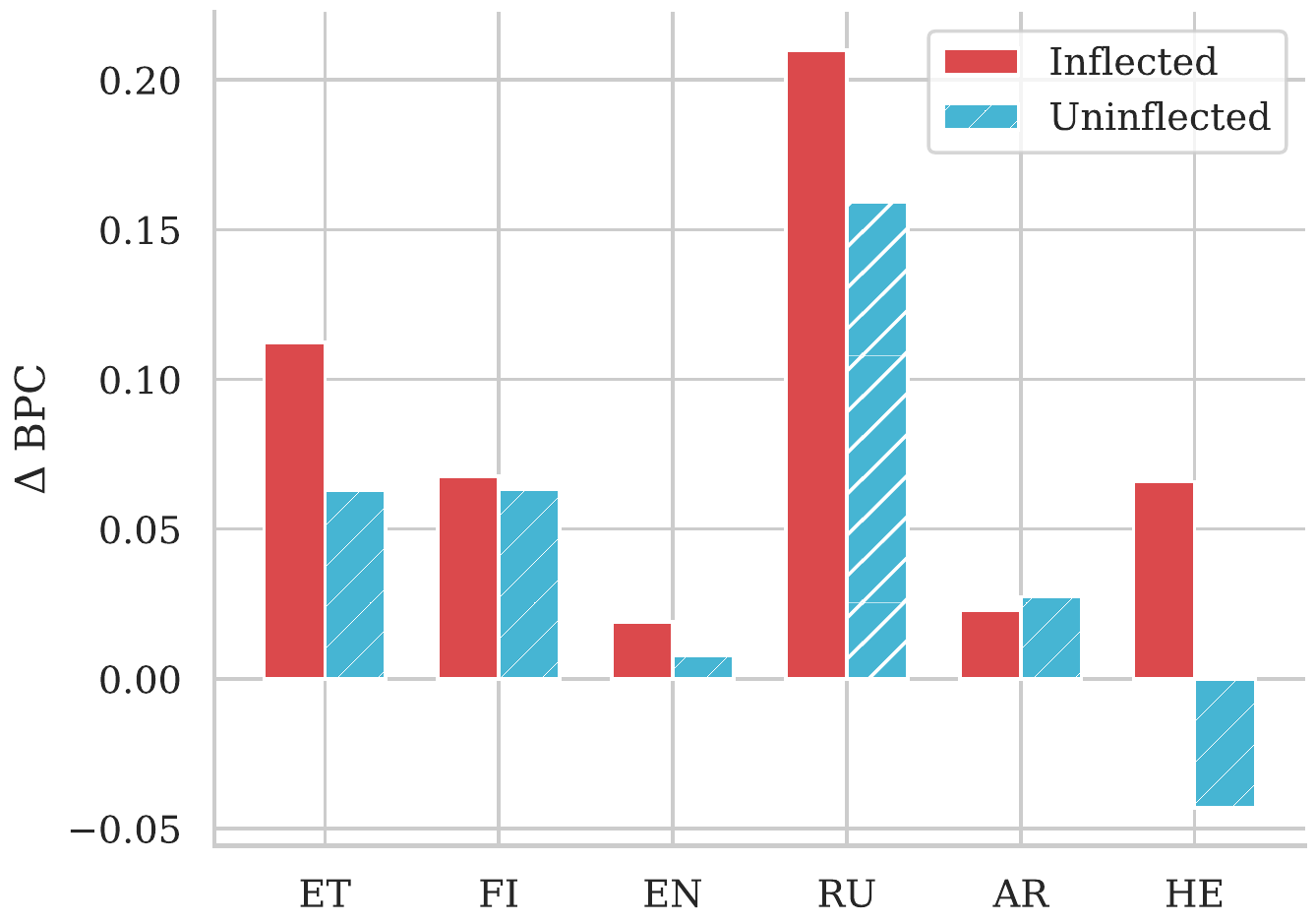}
    \label{infl-plot}}
\caption{Improvement of MTL over the LM baseline (a) over the inflection rate of each UD language, (b) over the quantity of training data for each UD language, and (c) for inflected and uninflected words in the UD dev set.}
\end{figure*}

\section{Language Modeling Results}
\paragraph{Distant MTL}\label{sec:mwc-results}
We first train CLMs where the language modeling data (from MWC) and morphology data (from UD) do not overlap (Table \ref{mwc-results-table}).\footnote{Since both of these datasets draw from Wikipedia, we verified that no sentences overlap between the MWC test set and the UD treebanks for each of the seven languages.} In this setting, we only train on the morphology features from UD and do not include this data as additional language modeling supervision. These models are trained on alternating batches from the two disjoint datasets. LM is a language modeling baseline with no multitask objective; MTL adds morphology supervision. 

We find that for all seven languages, the MTL model outperforms our baseline trained only on MWC. Our model also outperforms the strongest model from \citet{kawakami2017}, HCLMcache, which is a hierarchical language model with caching. Thus, adding morphology supervision to our character language models allows us to achieve lower BPCs than a more complicated LM architecture. 
Surprisingly, we see a larger gain on languages with more LM data (EN, DE, ES, FR) than those with less data (but are considered to be more morphologically rich, e.g., CS, DE, and RU); we explore this phenomenon more in Section \ref{sec:analysis}.

\begin{table*}
\subtable[]{
    \scriptsize
    \noindent\begin{tabular}{c|cc|ccc}
        \toprule
        & \textbf{\% Train} & \textbf{\# Chars} & \textbf{LM} & \textbf{MTL} & $\Delta$ \\
        \hline
        \multirow{6}{*}{\textbf{CS}} & 5\% & 0.31M & 2.829 & 2.793 & 0.036 \\
        & 10\% & 0.61M & 2.625 & 2.581 & 0.044 \\
        & 25\% & 1.5M & 2.379 & 2.303 & 0.076 \\
        & 50\% & 3.1M & 2.191 & 2.120 & 0.071 \\
        & 100\% & 6.1M & 2.013 & 1.938 & 0.075\\
        \cline{2-6}
        & MWC-LG & 10.2M & 1.835 & 1.729 & 0.106 \\
        \hline 
        \hline 
        \multirow{6}{*}{\textbf{RU}} & 5\% & 0.47M & 2.492 & 2.486 & 0.006 \\
        & 10\% & 0.93M & 2.305 & 2.283 & 0.022 \\
        & 25\% & 2.3M & 2.066 & 2.033 & 0.033 \\
        & 50\% & 4.7M & 1.935 & 1.898 & 0.037 \\
        & 100\% & 9.3M & 1.745 & 1.645 & 0.100 \\
        \cline{2-6}
        & MWC-LG & 18.2M & 1.554 & 1.377 & 0.177 \\
        \toprule
    \end{tabular}
    \label{vary-lm-table}}
\subtable[]{
    \scriptsize
    \noindent\begin{tabular}{c |c c|c c}
        \toprule
        & \textbf{\% Train} & \textbf{\# Chars} &  \textbf{BPC} \\
        \hline
        \multirow{5}{*}{\textbf{CS}} & 0\%  & (LM) & 2.013 \\
        \cline{2-4}
        &5\% & 0.34M & 2.040 \\
        & 10\% & 0.69M & 2.019 \\
        & 25\% & 1.7M & 2.000 \\
        & 50\% & 3.4M & 1.984 \\
        & 100\% & 6.9M & \textbf{1.938} \\
        \hline
        \hline
        \multirow{5}{*}{\textbf{RU}} & 0\%  & (LM) & 1.745 \\
        \cline{2-4}
        & 5\% & 0.27M & 1.758 \\
        & 10\% & 0.53M & 1.761 \\
        & 25\% & 1.3M & 1.673 \\
        & 50\% & 2.7M & 1.700 \\
        & 100\% & 5.3M & \textbf{1.645} \\
        \toprule
    \end{tabular}
    \label{vary-morph-table}}
\subtable[]{
    \scriptsize
    \noindent\begin{tabular}{c c|c}
    \toprule
        \textbf{LM data} & \textbf{Morph. data} & \textbf{BPC} \\
        \hline
        \multirow{4}{*}{\textbf{SK}} & None & 2.806 \\
        & SK & 2.779 \\
        & CS & 2.752 \\
        & CS+SK & 2.777 \\
        \hline
        \multirow{2}{*}{\textbf{CS+SK}} & None & 2.668 \\
        & CS+SK & \textbf{2.446} \\
        \hline
        \hline
        \multirow{4}{*}{\textbf{UK}} & None & 2.369 \\
        & UK & 2.348 \\
        & RU & 2.348 \\
        & RU+UK & 2.351 \\
        \hline
        \multirow{2}{*}{\textbf{RU+UK}} & None & 2.495 \\
        & RU+UK & \textbf{2.316} \\
    \toprule 
    \end{tabular}
    \label{low-resource-table}}
    \caption{(a) BPC on MWC development set with varied amounts of LM training data from MWC. The last line is from training on  MWC-large dataset, (b) BPC on MWC development set with varied amounts of supervised morphology data from UD train set (compared against the baseline LM), and (c) Cross-lingual transfer on UD, evaluated on low-resource language's development set: from Czech (CS; 6.9M characters in training set) to Slovak (SK; 0.4M) and from Russian (RU; 5.3M) to Ukrainian (UK; 0.5M)}
\end{table*}

\paragraph{Fully Supervised MTL}\label{sec:ud-results}
We then train CLMs using UD for both langauge modeling and morphology supervision on more languages (Table \ref{ud-results-table}). We again find that adding morphology supervision improves BPC. In general, we see smaller improvements between the LM and MTL models than under distant supervision, even though the UD LM data is fully annotated with morphology tags; this is likely due to the smaller training sets in UD (on average) than in MWC. On languages where the size of the two datasets are comparable, such as Russian and Czech, we see larger improvements on the fully supervised models than we do in the distant LM setting.  

To investigate these results, we compare the rate of inflected words on the development set (which we use as a rough measure of morphological complexity of the language) in a language against BPC improvement by MTL model (Fig. \ref{ud-plot-infl}). The rate at which each language is inflected is given in Table \ref{ud-results-table}. 
We unexpectedly find that how much a language benefits from morphology supervision is only weakly correlated with the inflection rate of the language (r=0.15). This is surprising, because one would expect that additional morphological supervision would help languages that encode more morphological features in the forms (i.e., with higher inflection rates).

We then examine the effect of training dataset size on BPC improvement between the LM and the multitasked model (Fig. \ref{ud-plot-train}). We find that more training data (which adds both morphological and LM supervision) is strongly correlated with larger gains over the baseline LM (r=0.93). Therefore, it seems that any potential correlation between morphological complexity and the benefit of multitasking morphology is overwhelmed by differences in dataset size. 

\section{Analysis Experiments}\label{sec:analysis}
\paragraph{Modeling Inflected Words}
We hypothesized that morphology supervision would be most beneficial to words whose form is dependent on their morphology, e.g. inflected words. To investigate this, we calculate BPC of our UD models on inflected and uninflected forms in the UD development set. We determine whether or not a word is inflected by comparing it to the (annotated) lemma given in the UD treebank. We find that on 16 of the 24 languages for which we train models on UD, the MTL model improves more on inflected words than uninflected words, and that the average delta between LM and MTL models is 31\% greater for inflected words than uninflected. A comparison of the improvements in six of these languages are given in Fig. \ref{infl-plot}. We show results for the agglutinative (ET, FI) and introflexive (AR, HE) languages and pick two fusional languages (EN, RU) against which to compare.

\paragraph{Effect of  Training Data}
One caveat to the observed gain from morphology is that the CLMs may capture this information if given more language modeling data, which is much cheaper to obtain than morphology annotations. To test this, we train CLMs on Czech (CS) and Russian (RU) on varied amounts of language modeling data from the MWC corpus (Table \ref{vary-lm-table}). We find that for both RU and CS, increasing the amount of LM data does not eliminate the gains we see from multitasking with morphology. Instead, we see that increasing LM data leads to \textit{larger} improvements in the MTL model. Even when we train the CLMs on twice as much LM data (obtained from a larger version of the MWC dataset, MWC-large), we continue to see large improvements via multitasking.

We then investigate how the amount of annotated morphology data affects performance on these models (Table \ref{vary-morph-table}). We find that, as expected, increasing the amount of  morphological data the language model is trained on improves BPC performance. For both Czech and Russian, the MTL models mulitasked with 25\% or more of the annotated  data still outperform the LM baseline, but MTL models trained on smaller subsets of the morphology data performed worse than the baseline. This is in line with our findings in Section \ref{sec:ud-results} that the amount of annotated morphology data is closely tied with how much multitasking helps. 

\paragraph{Cross-lingual Transfer}
In the previous section, we showed that the amount of training data (both for LM and for morphology) the CLM sees is crucial for better performance. Motivated by this, we extend our models to the cross-lingual setting, in which we use data from high-resource languages to improve performance on closely-related, low-resource ones. We train models on the (high, low) language pairs of (Russian, Ukrainian) and (Czech, Slovak) and transfer both LM and morphological supervision (Table \ref{low-resource-table}). We find the best performance for each low-resource language is achieved by using both the high-resource LM data and morphology annotations to augment the low-resource data. In Slovak (SK), this gives us a 0.333  BPC improvement over the MTL model on SK data alone, and in Ukranian (UK), we see a improvement of 0.032 in this setting over the MTL trained only on UK.

\section{Related Work}
Prior work has investigated to what degree neural models capture morphology when trained on language modeling \cite{vania2017} and on machine translation \cite{belinkov2017,bisazza2018}. Other work has looked into how the architecture of language models can be improved for morphologically-rich languages \cite{gerz2018}. In particular, both \citet{kawakami2017} and \citet{mielke2019} proposed hybrid open-vocabulary LM architectures to deal with rare words in morphologically-rich languages on MWC.\footnote{Results comparing against \citet{mielke2019} are given in the supplement, due to a different character vocabulary from \citet{kawakami2017}.}

Another line of work has investigated the use of morphology to improve models trained on other NLP tasks. These approaches add morphology as an input to the model, either with gold labels on the LM dataset \cite{vania2017} or by labeling the data with a pretrained morphological tagger \cite{botha2014,matthews2018}. This approach to adding morphology as input features to models has also been applied to dependency parsers \cite{vania2018} and semantic role labeling models \cite{sahin2018}. Unlike these approaches, however, our technique does not require the morphology data to overlap with the training data of the primary task or depend on automatically labeled features. More similarly to our work, \citet{dalvi2017} find that incorporating morphological supervision into the decoder of an NMT system via multitasking improves performance by up to 0.58 BLEU points over the baseline for English-German, English-Czech, and German-English.

\section{Conclusion}
We incorporate morphological supervision into character language models via multitask learning and find that this addition improves BPC on 24 languages. Furthermore, we observe this gain even when the morphological annotations and language modeling data are disjoint, providing a simple way to improve language modelsing without requiring additional annotation efforts. Our analysis finds that the addition of morphology benefits inflected forms more than uninflected forms and that training our CLMs on additional language modeling data does not diminish these gains in BPC. Finally, we show that these gains can also be projected across closely related languages by sharing morphological annotations. We conclude that this multitasking approach helps the CLMs capture morphology better than the LM objective alone. 

\section*{Acknowledgements}
This material is based upon work supported by the National Science Foundation Graduate Research Fellowship Program under Grant No. DGE-1762114.
We thank Victor Zhong, Sewon Min, and the
anonymous reviewers for their helpful comments.

\bibliography{acl2019}
\bibliographystyle{acl_natbib}

\appendix

\begin{table}[b!]
    \resizebox{\columnwidth}{!}{
    \begin{tabular}{c c c | c c c}
        \toprule
        & & & \multicolumn{3}{c}{\textbf{Num Chars}} \\
        \textbf{Lang} & \textbf{ISO} & \textbf{Treebank} & \textbf{Train} & \textbf{Dev} & \textbf{Test}  \\
        \hline
        Bulgarian & BG & BTB & 0.7M & 90K & 88K \\
        Catalan & CA & AnCora & 2.2M & 0.3M & 0.3M \\
        Czech & CS & PDT & 6.9M & 0.9M & 1.0M \\
        Danish & DA & DDT & 0.4M & 56K & 54K \\
        German & DE & GSD & 1.6M & 75K & 0.1M \\
        English & EN & EWT & 1.0M & 0.1M & 0.1M \\
        Spanish & ES & GSD & 2.1M & 0.2M & 65K \\
        Farsi & FA & Seraji & 0.6M & 77K & 78K \\
        French & FR & GSD & 1.9M & 0.2M & 52K \\
        Hindi & HI & HDTB & 1.3M & 0.2M & 0.2M\\
        Croatian & HR & SET & 0.9M & 0.1M & 0.1M \\
        Italian & IT & ISDT & 1.6M & 67K & 59K \\
        Latvian & LV & LVTB & 0.7M & 0.1M& 0.1M \\
        Dutch & NL & Alpino & 1.1M & 65K & 67K \\
        Polish & PL & LFG & 0.6M & 74K & 74K \\
        Portuguese & PT & Bosque & 1.1M & 58K & 55K \\
        Romanian & RO & RRT & 1.1M & 98K & 93K \\
        Russian & RU & SynTagRus & 5.3M & 0.7M & 0.7M \\
        Slovak & SK & SNK & 0.4M & 76K & 80K \\
        Ukranian & UK & IU & 0.5M & 71K & 98K \\
        \hline
        Estonian & ET & EDT & 2.2M & 0.2M & 0.3M \\
        Finnish & FI & TDT & 1.2M & 0.1M & 0.2M \\
        \hline 
        Arabic & AR & PADT & 1.3M & 0.2M & 0.2M \\
        Hebrew & HE & HTB & 0.8M & 63K & 68K \\
    \toprule
    \end{tabular}}
    \caption{Dataset statistics for Universal Dependencies (UD; v.2.3). Languages are grouped by typology, from top to bottom: fusional, agglutinative, and introflexive}
    \label{ud-statistics}
\end{table}

\section{Appendix: Languages and Datasets}
\label{sec:sup-datasets}
The languages we use from Universal Dependencies and details about their treebanks are given in Table \ref{ud-statistics}.  Most of the treebanks we used in this paper are manually annotated (and then possibly automatically converted to their current format), except for German, English, and French, which are automatically annotated. For models trained in the fully-supervised MTL setting where UD is used for both LM and morphology supervision, we calculate the character vocabulary for each language by including any character that occurs more than 5 times in the training set of the language's UD treebank.

Dataset statistics for the Multilingual Wikipedia Corpus (MWC) are given in Table \ref{mwc-statistics}. When analyzing the effect of LM training dataset size on Czech and Russian, we also train models on the training portion of a larger version of the MWC corpus, MWC-large, which contains approximately twice as much training data as the standard MWC dataset. Specifically, MWC-large contains 10.2M training characters for Czech and 18.2M for Russian. There is no prior work that we know of that reports BPC on this larger dataset.

\begin{table}
    \centering
    \small
    \begin{tabular}{c | c c c c}
        \toprule
        &  \multicolumn{3}{c}{\textbf{Num Chars}} \\
        \textbf{Lang} & \textbf{Vocab} & \textbf{Train} & \textbf{Dev} & \textbf{Test}  \\
        \hline
        CS & 238 & 6.1M & 0.4M & 0.5M \\
        DE & 298 & 13.6M & 1.2M & 1.3M \\
        EN & 307 & 15.6M & 1.5M & 1.3M \\
        ES & 307 & 11.0M & 1.0M & 1.3M \\
        FI & 246 & 6.4M & 0.7M & 0.6M \\
        FR & 272 & 12.4M & 1.3M & 1.6M \\
        RU & 273 & 9.3M & 1.0M & 0.9M \\
        \toprule
    \end{tabular}
    \caption{Dataset statistics for Multilingual Wikipedia Corpus (MWC). Vocabulary size is based on the character vocabulary given in \cite{kawakami2017}.}
    \label{mwc-statistics}
\end{table}

For models trained on the disjoint supervision setting, we use the character vocabulary provided for each language in the MWC dataset (see \citet{kawakami2017} for preprocessing details). In cases where we use two sources of supervision for the model -- LM supervision from MWC and morphology supervision from UD -- we use the MWC character vocabulary for all inputs, so that BPC results across models are comparable. This only affects a small number of the character types (11 or fewer for each language) in the UD training data. 

\begin{table*}
    \resizebox{\textwidth}{!}{
    \begin{tabular}{ c c c c c c c c c c c c c c c}
    \toprule
    & \multicolumn{2}{c}{\textbf{CS}} & \multicolumn{2}{c}{\textbf{DE}} & \multicolumn{2}{c}{\textbf{EN}} & \multicolumn{2}{c}{\textbf{ES}} & \multicolumn{2}{c}{\textbf{FI}} & \multicolumn{2}{c}{\textbf{FR}} & \multicolumn{2}{c}{\textbf{RU}} \\ 
    \hline
    & dev & test & dev & test & dev & test & dev & test & dev & test & dev & test & dev & test \\
    \hline 
    \textbf{Mielke BPE} & 1.88 & 1.856 & 1.45 & \textbf{1.414} & 1.45 & 1.386 & 1.42 & 1.362 & 1.70 & \textbf{1.652} & 1.36 & 1.317 & 1.63 & \textbf{1.598} \\
    \textbf{Mielke Full} & 1.95 & 1.928 & 1.51 & 1.465 & 1.45 & 1.387 & 1.42 & 1.363 & 1.79 & 1.751 & 1.36 & 1.319 & 1.74 & 1.709 \\
     \textbf{LM} & 2.01 & 1.975 & 1.52 & 1.493 & 1.45 & 1.395 & 1.55 & 1.482 & 1.74 & 1.705 & 1.60 & 1.565 & 1.72 & 1.692 \\
     \textbf{MTL} & \textbf{1.81} & \textbf{1.771} & \textbf{1.43} & \textbf{1.414} & \textbf{1.32} & \textbf{1.262} & \textbf{1.33} & \textbf{1.268} & \textbf{1.69} & 1.658 & \textbf{1.15} & \textbf{1.104} & \textbf{1.62} & \textbf{1.596} \\
    \toprule
    \end{tabular}}
    \caption{Results on Multilingual Wikipedia Corpus (MWC) in bits per character (BPC), trained on the vocabulary from \citet{mielke2019}.}
    \label{mwc-spell-table}
\end{table*}

The character vocabulary provided in the MWC dataset and used for the distant supervision setting differs from the vocabulary calculated by including the characters that occur more than 25 times in the MWC training set.\footnote{On English, this preprocessing difference decreases the character vocabulary size from 307 in the provided vocabulary to 167.} Because of this, our distant supervision setting on MWC is not comparable with \citet{mielke2019}, which uses the second vocabulary setting. Therefore, we retrain our character LM baselines and multitasked models in this vocabulary setting (Table \ref{mwc-spell-table}). We find that our LM and MTL models generally obtain slightly better performance on this setting, and we continue to see improvement from multitasking morphology over the character LM baseline.

\section{Appendix: Model Parameters and Training}
\label{sec:sup-params}
To train all models presented in this paper, we use the Adam optimizer \cite{kingma2015} with an initial learning rate of 0.002 and clip the norm of the gradient to 5. We also apply a dropout of 0.5 to each layer. We train each model on sequences of 150 characters and use early stopping with a patience of 10. We only use the language modeling performance (BPC) on the development set for early stopping and hyperparameter selection (and do not consider the morphology losses). For the UD language models, we train models with two hidden layers for 150 epochs with a batch size of 10. The models trained on MWC contain three hidden layers and are trained for 250 epochs with a batch size of 32. All of our models are implemented in Pytorch.\footnote{https://pytorch.org/} \\

\begin{table}
    \centering
    \small
    \begin{tabular}{c | c c | c c}
        \toprule
        & \multicolumn{2}{c|}{\textbf{Distant MTL}} & \multicolumn{2}{c}{\textbf{Fully-Supervised}} \\
        \textbf{Lang} & \textbf{MTL layer} & $\delta$ & \textbf{MTL layer} & $\delta$ \\
        \hline
        BG & - & - & 2 & 1.0 \\
        CA & - & - & 1 & 2.0 \\
        CS & 2 & 1.5 & 2 & 1.5\\
        DA & - & - & 2 & 0.5 \\
        DE & 2 & 2.0 & 1 & 1.0 \\
        EN & 2 & 1.0 & 2 & 1.0 \\
        ES & 2 & 1.0 & 1 & 1.5 \\
        FA & - & - & 1 & 0.01  \\
        FR & 3 & 1.0 & 1 & 2.0 \\
        HI & - & - & 1 & 2.0 \\
        HR & - & - & 2 & 1.0 \\
        IT & - & - & 1 & 2.0 \\
        LV & - & - & 2 & 1.0 \\
        NL & - & - & 2 & 1.5 \\
        PL & - & - & 2 & 1.0 \\
        PT & - & - & 2 & 1.5 \\
        RO & - & - & 2 & 0.5 \\
        RU & 3 & 1.0 & 2 & 1.5 \\
        SK & - & - & 2 & 2.0 \\
        UK & - & - & 2 & 1.0 \\
        \hline
        ET & - & - & 2 & 2.0 \\
        FI & 1 & 0.5 & 2 & 1.0 \\
        \hline 
        AR & - & - & 2 & 0.5 \\
        HE & - & - & 2 & 0.5 \\
        \toprule
    \end{tabular}
    \caption{Language specific parameters for multitasked models trained in the distant MTL setting and the fully-supervised MTL setting}
    \label{lang-params}
\end{table}

For each language, we individually tuned the level at which we multitask the morphology objectives and the weighting ratio between the primary and auxiliary losses $\delta$. We consider multitasking the morphology objective at either the first or second hidden layer (as all of our models have two hidden layers), and tune for each language $\delta = \{0.01, 0.1, 0.5, 1, 1.5, 2\}$. The parameters chosen for each language and setting (fully supervised or distant MTL) are given in Table \ref{lang-params}.
\\

\section{Appendix: Additional Results}
We provide the full set of results for our experiments in Section \ref{sec:analysis} on how well our CLMs model inflected forms versus uninflected forms across all 24 UD languages (Table \ref{infl-analysis-full}).

\begin{table}
    \small
    \centering
    \begin{tabular}{c c c|c c c}
        \toprule
        \textbf{Lang} & \textbf{\%Infl} & \textbf{Word Type} & \textbf{LM} & \textbf{MTL} & $\Delta$ \\
        \hline
        \multirow{2}{*}{BG} & \multirow{2}{*}{39\%} & inflected & 2.092 & 2.085 & \textbf{0.008}\\
        & & uninflected & 2.333 & 2.330 & 0.002\\
        \hline
        \multirow{2}{*}{CA} & \multirow{2}{*}{31\%} & inflected & 1.849 & 1.783 & \textbf{0.066}\\
        & & uninflected & 2.007 & 1.943 & 0.064\\
        \hline
        \multirow{2}{*}{CS} & \multirow{2}{*}{43\%} & inflected & 2.205 & 1.940 & \textbf{0.265}\\
        & & uninflected & 2.539 & 2.322 & 0.217\\
        \hline
        \multirow{2}{*}{DA} & \multirow{2}{*}{30\%} & inflected & 2.411 & 2.387 & \textbf{0.024}\\
        & & uninflected & 2.559 & 2.552 & 0.007\\
        \hline
        \multirow{2}{*}{DE} & \multirow{2}{*}{33\%} & inflected & 1.916 & 1.868 & 0.048\\
        & & uninflected & 2.323 & 2.263 & \textbf{0.060}\\
        \hline
        \multirow{2}{*}{EN} & \multirow{2}{*}{15\%} & inflected & 2.235 & 2.216 & \textbf{0.019}\\
        & & uninflected & 2.579 & 2.571 & 0.008\\
        \hline
        \multirow{2}{*}{ES} & \multirow{2}{*}{28\%} & inflected & 1.742 & 1.700 & 0.042\\
        & & uninflected & 2.053 & 2.010 & \textbf{0.043}\\
        \hline
        \multirow{2}{*}{FA} & \multirow{2}{*}{27\%} & inflected & 2.874 & 2.859 & \textbf{0.016}\\
        & & uninflected & 2.499 & 2.492 & 0.007 \\
        \hline
        \multirow{2}{*}{FR} & \multirow{2}{*}{32\%} & inflected & 1.856 & 1.809 & 0.047\\
        & & uninflected & 2.228 & 2.174 & \textbf{0.054}\\
        \hline
        \multirow{2}{*}{HI} & \multirow{2}{*}{28\%} & inflected & 1.996 & 1.941 & \textbf{0.053}\\
        & & uninflected & 2.270 & 2.228 & 0.042 \\
        \hline
        \multirow{2}{*}{HR} & \multirow{2}{*}{49\%} & inflected & 2.055 & 2.021 & \textbf{0.035} \\
        & & uninflected & 2.507 & 2.487 & 0.021\\
        \hline
        \multirow{2}{*}{IT} & \multirow{2}{*}{36\%} & inflected & 1.897 & 1.852 & 0.045\\
        & & uninflected & 2.056 & 2.010 & \textbf{0.046}\\
        \hline
        \multirow{2}{*}{LV} & \multirow{2}{*}{47\%} & inflected & 2.387 & 2.361 & \textbf{0.027} \\
        & & uninflected & 2.782 & 2.758 & 0.024 \\
        \hline
        \multirow{2}{*}{NL} & \multirow{2}{*}{19\%} & inflected & 2.161 & 2.493 & \textbf{0.030}\\
        & & uninflected & 2.131 & 2.468 & 0.025 \\
        \hline
        \multirow{2}{*}{PL} & \multirow{2}{*}{42\%} & inflected & 2.522 & 2.462 & \textbf{0.060} \\
        & & uninflected & 2.633 & 2.578 & 0.054 \\
        \hline
        \multirow{2}{*}{PT} & \multirow{2}{*}{31\%} & inflected & 2.071 & 2.065 & 0.007 \\
        & & uninflected & 2.214 & 2.205 & \textbf{0.009} \\
        \hline
        \multirow{2}{*}{RO} & \multirow{2}{*}{42\%} & inflected & 2.037 & 1.987 & 0.050 \\
        & & uninflected & 2.373 & 2.316 & \textbf{0.057} \\
        \hline
        \multirow{2}{*}{RU} & \multirow{2}{*}{42\%} & inflected & 2.130 & 1.920 & \textbf{0.210} \\
        & & uninflected & 2.583 & 2.424 & 0.159 \\
        \hline
        \multirow{2}{*}{SK} & \multirow{2}{*}{45\%} & inflected & 2.976 & 2.969 & 0.007 \\
        & & uninflected & 3.545 & 3.535 & \textbf{0.010} \\
        \hline
        \multirow{2}{*}{UK} & \multirow{2}{*}{40\%} & inflected & 2.580 & 2.553 & \textbf{0.027} \\
        & & uninflected & 2.553 & 2.956 & 0.009 \\
        \hline
        \hline
        \multirow{2}{*}{ET} & \multirow{2}{*}{49\%} & inflected & 2.397 & 2.692 & \textbf{0.112}\\
        & & uninflected & 2.285 & 2.629 & 0.063\\
        \hline
        \multirow{2}{*}{FI} & \multirow{2}{*}{55\%} & inflected & 2.152 & 2.084 & \textbf{0.068}\\
        & & uninflected & 2.402 & 2.339 & 0.063\\
        \hline 
        \hline
        \multirow{2}{*}{AR} & \multirow{2}{*}{86\%} & inflected & 2.036 & 2.013 & 0.023 \\
        & & uninflected & 3.856 & 3.828 & \textbf{0.027}\\
        \hline
        \multirow{2}{*}{HE} & \multirow{2}{*}{42\%} & inflected & 3.426 & 3.360 & \textbf{0.066}\\
        & & uninflected & 2.168 & 2.211 & -0.043\\
        \toprule
    \end{tabular}
    \caption{BPC performance on the UD development set on inflected versus uninflected words. Bold delta values for each language indicate whether than language improves more on inflected or uninflected words by when multitasking morphology is added.}
    \label{infl-analysis-full}
\end{table}

\end{document}